\documentclass[12pt,letterpaper]{article}
\usepackage{standalone}
\usepackage[sort&compress,super]{natbib}
\usepackage[british,french]{babel}
\babelprovide[import,main]{english}
\usepackage{newunicodechar} 
\usepackage[dvipsnames,svgnames]{xcolor}
\usepackage[hyphens]{url}
\usepackage{amsmath}
\usepackage{listings}

\usepackage{float}
\usepackage{tocloft}
\usepackage{ifthen}
\usepackage[utf8]{inputenc}
\usepackage[T1]{fontenc}
\usepackage{lmodern} 
\usepackage{eso-pic}
\usepackage{enumerate}
\usepackage{paralist}
\usepackage{setspace}
\usepackage{longtable}
\usepackage{tabularx}

\usepackage{multirow}
\usepackage{array}
\newcolumntype{H}{>{\setbox0=\hbox\bgroup}l<{\egroup}@{}}
\usepackage{float}
\usepackage[figuresright]{rotating}
\usepackage{titleps}
\usepackage{filecontents}
\usepackage{bibentry}

\usepackage{pdfpages}
\usepackage{pdflscape}
\usepackage[fencedCode,hybrid]{markdown}
\usepackage{changepage}
\usepackage{randtext} 
\usepackage{graphicx}
\usepackage{pgf}
\usepackage{tikz}
\usepackage{soul}
\usepackage[top=2.54cm,bottom=1.8cm,left=3.17cm,right=3.17cm]{geometry}
\usepackage[format=plain,justification=justified,font=small,skip=6pt]{caption}
\bibpunct{(}{)}{;}{a}{,}{,}

\renewcommand{\refname}{\normalsize References}
\newsavebox{\boxtext}

\makeatletter
\newcommand*{\textlabel}[2]{%
  \edef\@currentlabel{#1}
  \phantomsection
  #1\label{#2}
}
\makeatother

\newunicodechar{ﬀ}{ff} 
\newunicodechar{ﬁ}{fi} 
\newunicodechar{ﬂ}{fl}
\newunicodechar{ﬃ}{ffi}
\newunicodechar{ﬄ}{ffl}
\newunicodechar{ }{\\ \space\\}

\def\thisMonth{%
  \ifcase\month\or January\or February\or March\or  April\or May\or June\or July\or August\or September\or October\or November\or December\fi%
  \space\number\year
  }
\def\toDay{\number\day\space\thisMonth}

\def\thisMonthFr{%
  \ifcase\month\or janvier\or février\or mars\or avril\or mai\or juin\or juillet\or août\or septembre\or Octobre\or novembre\or décembre\fi%
  \space\number\year
  }

\begin{document}
\shorthandoff{:} 
\shorthandoff{;}
\shorthandoff{!}
\shorthandoff{?}


\def\writing{1}

\bibpunct{}{}{,}{s}{,}{,}
\renewcommand*{\thesection}{}
\def\sec#1{\section*{#1}
\vspace{-12pt}\addcontentsline{toc}{section}{#1}
}
\def\subsec#1{\subsection*{\normalfont{\textbf {#1}}\vspace{-6pt}}
\addcontentsline{toc}{subsection}{#1}
}
\def\subsubsec#1{\subsubsection*{\normalfont{\textit {#1}}\vspace{-6pt}}
\addcontentsline{toc}{subsubsection}{#1}
}
\renewcommand{\refname}{\\[-86pt]\section*{\textsc References\vspace{-6pt}}
\addcontentsline{toc}{section}{References}
}
\renewcommand{\labelitemi}{\textbullet}
\footskip = 18pt
\setlength{\parindent}{28pt}
\parskip 8pt

\def\agonist{agonist}
\def\anal{analysis}
\def\endo{endometriosis}
\def\Endo{Endometriosis}
\def\drug{\mbox{GnRH-a}}
\def\Drug{Gonadotropin-releasing hormone}
\def\ma{meta-analysis}
\def\MA{Meta-analysis}
\def\mesh{MeSH term}
\def\pico{PICO}
\def\prisma{PRISMA}

\def\quest#1{\textcolor{green}{#1}}
\def\rev#1{#1}
\def\excerpt#1#2{%
  \begin{quotation}
  \noindent {\it #1} (#2)
  \end{quotation}
}
\usetikzlibrary{arrows,backgrounds,calc,positioning,shadows,shapes,shadows.blur,shapes.symbols}
\tikzstyle{Boxed} = [draw=black!58, rectangle, minimum width=1mm, minimum height=1mm, font=\small, text opacity=1, top color=cyan!10, bottom color=cyan!10, fill opacity=1, very thick, drop shadow={blue!10, rounded corners=2pt, fill opacity=0.1, shade}]
\def\BackgroundPic#1{
\begin{tikzpicture}[remember picture, overlay]
\node[yshift=22cm] (uni) at (current page.south) 
        {\includegraphics[scale=0.6]{#1}}; 
\end{tikzpicture}
}

\def\discipline{\vspace*{180pt}Méthodologie et Statistiques\\en\\Recherche Biomédicale}
\def\topic{Computational-Assisted Systematic Review and Meta-Analysis (CASMA): Effect of a Subclass of GnRH-a on Endometriosis Recurrence
}
\def\shorttitle{Computational-Assisted SRMA (CASMA) - Endometriosis Recurrence}

\author{%
  {\large
  Sandro Tsang, PhD}\\
  Faculty of Medicine\\
  Université Paris-Saclay\\
  \texttt{\url{@: skf.tsang[at]gmail.com}}\\
  \texttt{\url{https://santsang.github.io/}}\\
  \texttt{\url{https://orcid.org/0000-0002-5144-9049/}}\vspace{-12pt}
}
\date{{\normalsize \toDay}\vspace{-15pt}}
\title{\fontsize{18}{16}\selectfont{\topic}\vspace{-12pt}}
\maketitle
\makeatother
\thispagestyle{empty}







\begin{abstract}
{\fontsize{10.5}{10.5}\selectfont
\noindent \textbf{Background:} Medical literature continues to grow at an astonishing rate, rendering the traditionally manual, resource-intensive PRISMA guidelines difficult to uphold. This bottleneck risks methodological transparency and efficiency, demanding computational solutions.\\
\noindent \textbf{Objective:} Building on prior work in information retrieval from unstructured medical data, this study introduces a framework, Computational-Assisted Systematic Review and Meta-analysis (CASMA), to enhance the efficiency, transparency, and reproducibility of evidence synthesis. Endometriosis recurrence, marked by inconsistent definitions, serves as the ideal clinical case to demonstrate the framework's application.\\
\noindent \textbf{Methods:} The CASMA framework integrates standard PRISMA guidelines with fuzzy matching and regular expression (regex) search to facilitate deduplication and pre-screening of relevant records prior to manual paper selection.  A specific subclass of gonadotropin-releasing hormone agonists (referred to as \drug\ interchangeably) was chosen to avoid confounding from a potential response to an intrauterine device. A modified splitting method addressed unit-of-analysis errors in multi-arm trials, alongside other sensitivity, subgroup analyses, bias assessment, and GRADE.\\
\noindent \textbf{Results:} This semi-automated process of pre-screening sharply reduced the labour-intensive workflow: From fetching 33,444 records, the computational pre-screening identified 29 potentially eligible records (including systematic reviews and RCTs) for manual assessment in only 11 days. Hand searches were performed on six manually selected meta-analyses to identify seven eligible RCTs for evidence synthesis (841 patients; 152 recurrences). The pooled random-effects model yielded a statistically significant Risk Ratio (RR) of $0.64$ ($95\%$ confidence interval (CI) $(0.48$ to $0.86)$) (or a $36\%$ reduction in recurrence), with non-significant heterogeneity ($I^2=0.00\%$, $\tau^2=0.00$). The profiling likelihood method confirmed $\tau^2=0.00$, but its $95\%$ CI $(0.00$ to $0.59)$ indicates some level of uncertainty. Sensitivity and subgroup analyses supported the robustness and stability of the~findings.\\
\noindent \textbf{Conclusion:} The consistency of the results with the existing evidence reinforces that this subclass of \drug\ is a promising management strategy to reduce \endo\ recurrence. The CASMA framework is a validated, efficient, and reproducible approach to help deliver medicine backed by the current best evidence. This study bridges the gap between medical research and computer science, offering a generalisable solution for managing the rapidly growing medical literature.\\[6pt]
}
\end{abstract}

\onehalfspacing

\onehalfspacing
\pagenumbering{arabic}
\setcounter{page}{2}
\sec{Introduction}

Endometriosis has been documented in the medical literature since the mid-19th century, yet treating the disease remains a source of considerable clinical uncertainty.\cite{Koninckxetal2021,Tomassettieta2021} \Endo~is a common gynaecological condition affecting over 190 million women worldwide,\cite{Ellisetal2022} and it contributes to the Global Burden of Disease.\cite{Fengetal2022} While recurrence was once considered a rare event, it is now recognised as a common and complex challenge.\cite{Ceccaronietal2019} The rates of recurrence vary across the world, partially due to inconsistent definitions.\cite{Ceccaronietal2019} Consequently, the observed recurrence rate is a function of both biological persistence and methodological definition. Post-operative hormonal suppression is a widely used strategy to manage endometriosis and reduce the risk of recurrence.\cite{Zakharietal2021,Zhengetal2016} The existing evidence reports a wide range of effects for these agents, from no effect to a significant protective effect,\cite{Ceccaronietal2019,Chenetal2020,Zakharietal2021,Zhengetal2016} which makes translating evidence into clinical practice difficult. Searching for definitive, confirmatory evidence is therefore a critical research priority.

The immense and continuously expanding volume of literature on \endo~presents a substantial challenge to evidence synthesis when strictly compiling with manual, resource-intensive PRISMA guidelines. This is exemplified by the exponential growth in endometriosis publications since the 1960s (see \mbox{Figures~S.1 and S.2}). A Computational-Assisted Systematic Review and Meta-analysis (CASMA) framework was therefore employed to scale up evidence synthesis with enhanced transparency and reproducibility. Following the standard PRISMA guidelines, regular expressions (regex) search and fuzzy matching were introduced to pre-screen the records before a manual selection process, efficiently excluding 812 records in a few days. This paper applies the CASMA framework to synthesise evidence about a specific subclass of \drug~(or \drug~interchargeably) where the drug delivery does not involve the use of an intrauterine device (IUD). This arrangement avoids the confounding reaction to a foreign body, leading to clearer and more precise results for easier interpretation. The aim was to synthesise robust evidence on the efficacy of the \drug~subclass in reducing endometriosis recurrence by deploying the CASMA framework to achieve validated, reproducible, transparent, and efficient evidence synthesis that leverages expert knowledge effectively.

\sec{Methods}

\subsec{Protocol and Registration}
Prospective registration of the review protocol was not feasible due to time constraints. Instead, the protocol was retrospectively registered on the Open Science Framework (OSF); DOI: \url{https://doi.org/10.17605/OSF.IO/R2DFA}, registered \mbox{5 September 2025}.

\subsec{Search Strategy}

Following PRISMA guidelines,\cite{Higginsetal2019} a systematic search was conducted in PubMed, Scopus, Google Scholar, and CrossRef between 6 and 17 June 2025. A final free-text search was performed on 14 July 2025 after the research direction was refined. Search terms combined controlled vocabulary (e.g., MeSH) and free text to capture the population (“endometriosis”), interventions (“GnRH”, “hormonal therapy”), and study designs (“meta-analysis”, “controlled trial”, “clinical trial”), with adaptations for each database. The search terms were derived from several much cited reviews\cite{Koninckxetal2021,Kvaskoffetal2022,OlivePritts2001} and a post by Mayo Clinic.\cite{MayoClinic2025b} The full search syntax is available as Supplementary Material.

To enhance efficiency, accuracy, and reproducibility, semi-automated text-matching techniques -- including fuzzy matching and regular expressions (regex) -- were integrated into the PRISMA workflow. This computational approach, which facilitates a rapid assessment of records prior to manual screening, is inspired by a previously proposed method for analysing unstructured medical data.\cite{Tsang2020} Specifically, fuzzy matching was performed on the titles to deduplicate records (an example is shown in \mbox{Table~S.2} in the Supplementary section). The native \texttt{R}\cite{R2022} functions were implemented to identify MAs of RCTs and to exclude records concerning network meta-analyses, Bayesian studies, or associated diseases such as adenomyosis. An automated research tool\cite{ResearchRabbit2021} was employed to ensure that conducting an update of a meta-analysis was an appropriate research direction after the manual selection process was exhausted.

\subsec{Eligibility Criteria and Outcome Definition}

The initial plan was to conduct a meta meta-analyses, which was subsequently refined to a meta-analysis of randomised controlled trials (RCTs). Eligible studies were RCTs evaluating post-operative hormonal treatment for managing endometriosis, with a control arm receiving placebo, expectant management, or no treatment. Studies were excluded if the comparator (in)directly affected sex hormone levels or the hormonal therapy involved intrauterine devices (IUDs), as IUDs might introduce confounding due to the foreign body response.

The primary outcome measure was the recurrence of endometriosis. Leading medical associations agreed that recurrence could be diagnosed through four established methods: symptom-based, image-based, laparoscopic, or histological diagnosis.\cite{Tomassettieta2021} \Endo~is a complex disease, often presenting with various symptoms and frequently coexisting with adenomyosis, which is often mistakenly diagnosed as \endo.\cite{Oliveetal1993,Tomassettieta2021} Studies defining recurrence solely based on the resolution or re-emergence of individual symptoms alone were excluded to ensure that synthesised evidence about recurrence fitted the standardised terminology.\cite{Tomassettieta2021}




\subsec{Inter-rater Reliability (IRR)}

Reviewer agreement for selecting meta-analysis papers was analysed in two stages. A three-point grading system was employed to record the selection decision and for computing the IRRs after the entire review process. This approach helps to avoid selection bias. Discrepancies in paper selection were resolved through discussion. The statistical approach was based on non-parametric bootstrapping with 2\,000 replications.

For the title/abstract review, a concordance of 82.76\% (24/29 papers) was reached, yielding a weighted absolute Kappa ($\kappa$) of 0.68 (95\% bootstrap bias-corrected and accelerated (BCa) confidence interval (CI) \mbox{(0.396 to 0.884)}). The agreement is substantial.\cite{LandisKoch1977} For the full-text review, only slight agreement was achieved; the concordance was 57.14\% (8/14 papers), and the $\kappa$ was 0.05 (95\% bootstrap CI \mbox{(-0.286 to 0.588)}). A negative $\kappa$ indicates that the observed agreement was lower than expected by chance.\cite{LandisKoch1977} The true level of agreement could plausibly range from poor to moderate.\cite{LandisKoch1977,Vierkant1997}  Table~S.2 presents the grading at the meta-analysis selection stage, and the bootstrap results of each replication are depicted in \mbox{Figure~S.3(a) and S.3(b)}.

\subsec{Data Extraction}
ST extracted data on study design, patient characteristics, interventions, comparators, and recurrence outcomes, recording them in a structured spreadsheet that followed the format of well-written publications. A subset of data was independently extracted by another reviewer in the same approach. Discrepancies in data extraction and risk of bias (RoB) assessment were resolved by ST after cross-checking the data against published systematic reviews, including a Cochrane Systematic Review.

\subsec{Statistical Analysis}

Analyses were conducted on a device with a base clock speed of 2.55~GHz, 8~GB RAM, and 8 threads. \texttt{R} (version~4.2.2) was the main statistical implementation.\cite{R2022} \texttt{data.table}\cite{Barrettetal2025} and \texttt{Hmisc}\cite{Harrell2024} were the packages used for data wrangling. Inter-rater reliability was computed using the \texttt{psy}\cite{Falissard2020} package with weighted absolute Kappa. \texttt{boot}\cite{CantyRipley2024} was employed for all bootstrapped statistics. Effect estimates for \drug~were calculated with the \texttt{metafor} package,\cite{Viechtbauer2010} and risk-of-bias assessments were depicted using \texttt{robvis}.\cite{McguinnessHiggins2025} Risk ratios (RRs) were pooled using a random-effects model to account for between-trial variability. A splitting method was applied to adjust for multi-arm RCTs and mitigate unit-of-analysis errors.\cite{Axonetal2023} The method was modified by proportionally splitting the control group to match the size of the intervention arm of interest relative to that of all intervention arms. Truncation was applied to ensure integer counts. Cumulative recurrence curves were digitised when necessary to approximate intention-to-treat (ITT) counts.\cite{Drevonetal2017} When ITT statistics and a cumulative recurrence curve were not available, loss to follow-up was assumed to be independent from recurrence.

A non-parametric bootstrap was performed to ensure robustness. For the approximations of the IRRs at the two stages, 2\,000 replications were applied, and 10\,000 replications were applied to verify the stability of the pooled RR and its CI. The profile likelihood method was used to derive a robust 95\% CI for the heterogeneity parameter ($\tau^2$), with the estimation performed with 50 steps to ensure numerical stability.\cite{HardyThompson1996}

Other sensitivity and subgroup analyses were performed to check the stability and robustness of the primary model. The Cochrane RoB 2 tool was the basis for assessing the 5 domains of RoB of the included randomised controlled trials (RCTs).\cite{Sterneetal2019} Publication bias was assessed using a Doi plot and LFK index due to the small number of included studies ($n=7$).\cite{FuruyaKanamorietal2018} The certainty of evidence for the primary outcome was judged using the GRADE framework. The Summary of Findings table was obtained from GRADEpro.\cite{GRADEpro2025}

\sec{Results}

Our initial search identified 33\,444 endometriosis documents from four journal databases (278 were potentially meta-analyses of RCTs) (see \mbox{Figure~1}). Grey literature records were extracted from Google Scholar and Crossref (the three categories stated in lower panel of the \mbox{Figure~S.2}). The workflow followed the PRISMA procedure, but introduced a semi-automated pre-screening stage prior to manual screening. With the application of fuzzy matching and regex search, 745 records were eliminated. Combined with the two free-text search records, only 29 records required manual screening. The workflow, including record fetching 33\,444, took only 11 days to complete. This process ultimately yielded two eligible MAs with incomparable comparators. The decision to perform an update of an older MA\cite{Zhengetal2016} was supported by the finding of an automated research tool\cite{ResearchRabbit2021} (see \mbox{Figure~S.4}). The computation-assisted systematic review and meta-analysis (CASMA) process was not only efficient, but also reliable. Following data extraction and discussion, seven RCTs were confirmed as eligible for the evidence synthesis.

A summary of the included studies is presented in \mbox{Table~1}. Four trials were conducted in Italy, two in China, and one was in the US or Canada. In these studies, 443 patients received GnRH-a, and 398 received no treatment, a placebo, or expectant management. All but one trial\cite{Hornsteinetal1997} evaluated the risk of recurrence using GnRH-a depots. In one study,\cite{Huangetal2018} the specific drug was not reported; since it was administered subcutaneously, it was classified as a GnRH-a depot. Two trials assessed endometriosis recurrence based on symptoms.\cite{Hornsteinetal1997,Vercellinietal1999} Five trials followed patients for 2 years, while two had follow-up periods of 1.5 or 5 years. All participants were of reproductive age, but the accrual periods varied. The quality of research in two trials was undermined by not reporting the accrual period\cite{Hornsteinetal1997} or enrolling only 100 patients over 4 years.\cite{Huangetal2018}

\begin{figure}[H]
  \caption{PRISMA Flow Chart for Study Selection}
\begin{minipage}[t]{.99\textwidth}
    \centering
    \includegraphics[scale=0.63,trim = 15mm 43mm 5mm 20mm,clip]{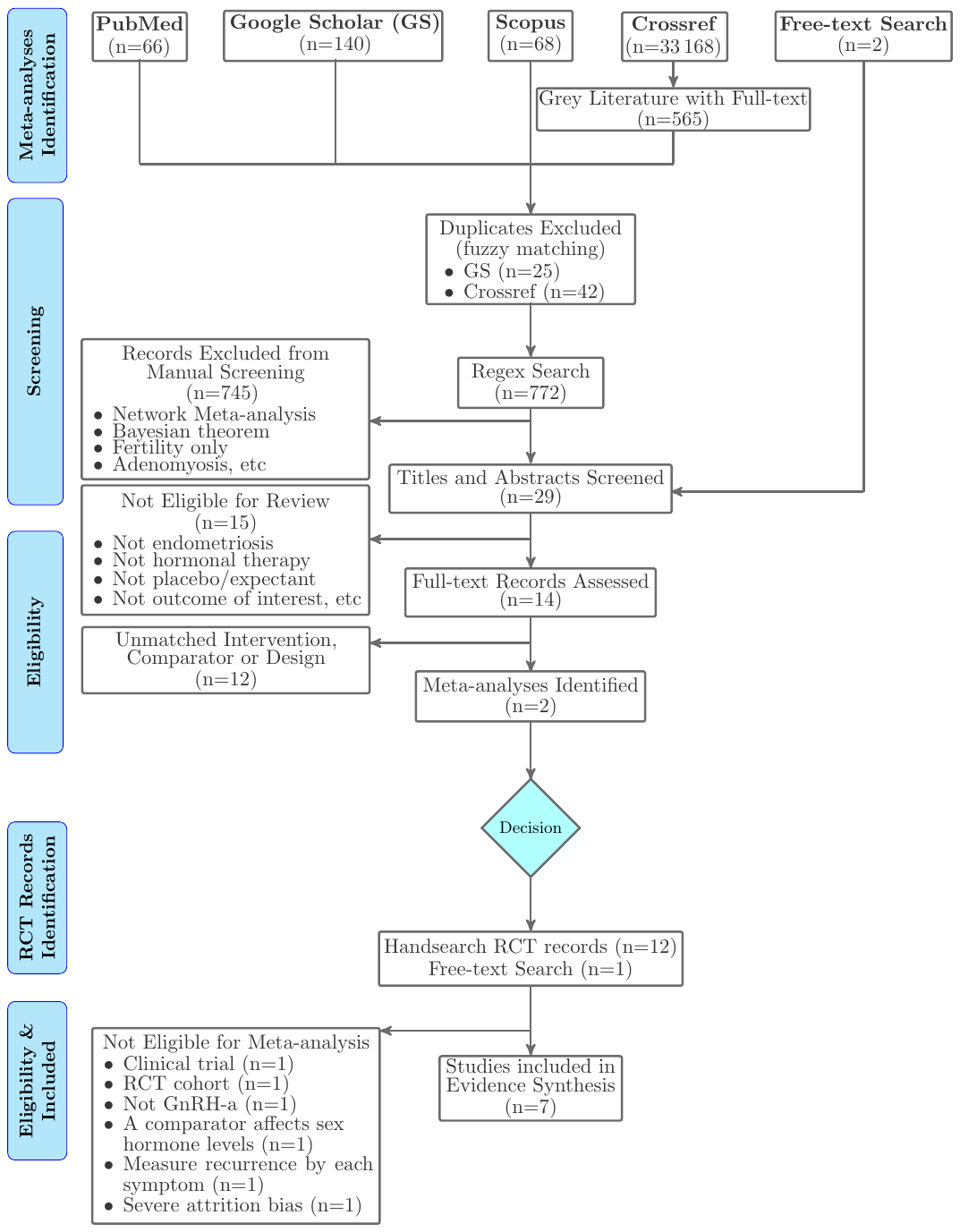}
\end{minipage}%
\end{figure}\vspace{-12pt}

The primary random-effects (RE) model (Figure~2(a)) pooled data from \mbox{7 papers}, including 841 patients and 152 inferred recurrences. The pooled risk ratio (RR) was 0.64 (95\% CI \mbox{(0.48 to 0.86)}). This was statistically significant as the 95\% CI excluded the null value (1). The risk reduction was 36\% ($=(1-0.64)\times100$) for patients who received this subclass of GnRH-a agents. This analysis included the 24-month recurrence from a trial that also reported a 12-month recurrence.\cite{Vercellinietal1999} When the analysis was performed with the 12-month outcome, the pooled effect reduced to \mbox{RR = 0.59} (95\% CI \mbox{(0.43 to 0.81)}), indicating a slightly stronger protective effect (see \mbox{Figure 3}). However, the differences between the two analyses were not practically meaningful, as the 95\% CIs overlapped substantially.

\begin{center}
  \vspace*{72mm}
  \newsavebox{\tablebox}  \setbox\tablebox=\vbox{\newcolumntype{P}[1]{>{\raggedright\arraybackslash}p{#1}}
\newcolumntype{C}[1]{>{\centering\arraybackslash}p{#1}}
\newcolumntype{H}{>{\setbox0=\hbox\bgroup}l<{\egroup}@{}}
\fontsize{7}{7.5}\selectfont
\renewcommand*{\arraystretch}{1.2}
\setlength{\tabcolsep}{1.5pt}
\begin{longtable}{
  l @{\space\space} 
  C{2cm} @{\space}
  P{1.5cm} 
  c 
  P{1.7cm} 
  P{1.8cm} 
  P{1.8cm} 
  P{1.8cm} 
  P{1.8cm} 
  c 
  c  
  P{1.5cm}  
  c  
  c  
  H 
}

\hline
\multicolumn{1}{l}{study}& 
\multicolumn{1}{c}{country}& 
\multicolumn{1}{c}{design}& 
\multicolumn{1}{c}{accrual}& 
\multicolumn{1}{c}{baseline}& 
\multicolumn{1}{c}{age}& 
\multicolumn{1}{c}{surgery}& 
\multicolumn{1}{c}{\multirow{2}{1.5cm}{recurrence diagnosis}}&
\multicolumn{1}{c}{intervention}& 
\multicolumn{1}{c}{\multirow{2}{1.5cm}{medication duration}}& 
\multicolumn{1}{c}{control}& 
\multicolumn{1}{c}{follow-up}& 
\multicolumn{2}{c}{Recurrence}\\
\cline{13-14}
 & & & & & & & & & & & & 
\multicolumn{1}{c}{GnRH-a}& 
\multicolumn{1}{c}{Control}\\ 
\hline
Hornstein et al. (1997)\cite{Hornsteinetal1997}&Canada \& US&Multicentre – 2 arms&unknown&stages II-IV&\mbox{I: 30.4±6.0;} \mbox{C: 31.1±6.2}&laser or electrosurgery&symptomatic -- composite measure&Nafarelin nasal 400µg&6 months&Placebo&24 months&17/56&26/53&The authors provide the As-treated (AT) recurrence statistics.  The intent-to-treat statistics were inferred by annotating the Kaplan Meier cumulative recurrence curve with WebPlotDigitizer to infer the probability.\tabularnewline
Vercellini et al. (1999)\cite{Vercellinietal1999}&Italy&Multicentre – 2 arms&02/92-06/94&stages I-IV&\mbox{I: 30.1±5.4;} \mbox{C: 30.0±5.3}&laparoscopy&symptomatic -- composite measure&Goseline SC 3.6mg&6 months&Expectant&24 months&23/133&32/134&AT recurrence statistics at 12 and 24 months were provided. The ITT statistics were inferred by annotating the Kaplan-Meier cumulative recurrence curve with WebPlotDigitizer to infer the probability.\tabularnewline
Busacca et al. (2001)\cite{Busaccaetal2001}&Italy&Multicentre - 2 arms&01/97-12/99&stages III-IV&\mbox{I: 20-37;} \mbox{C: 21-38}&laparoscopy&gynaecological exam and/or pelvic echography&Leuropelin acetate SC 3.75mg&3 months&Expectant&36 months&4/44&4/45&ITT statistics were reported.\tabularnewline
Loverro et al. (2008)\cite{Loverro2008}&Italy&2 arms&01/98-01/99&stages III-IV&\mbox{I: 28.7±4.4;} \mbox{C: 28.5±4.5}&laparoscopic&laparoscopic&Triptorelin depot 3.75mg&3 months&Placebo&60 months&4/30&2/30&We inferred the ITT with the assumption that the loss to follow up was not associated with endometriosis recurrence; i.e, best scenario was assumed.\tabularnewline
Sesti et al. (2009)\cite{Sestietal2009}&Italy&4 arms&01/04-08/06&stages III-IV&\mbox{I: 30.8±6.0;} \mbox{C: 31.3±5.1}&laparoscopic (uni-) or (bil-)ateral cystectomy&Echography and  Second-look laparoscopy&Group 2: Tryptorelin or Leuropelin 3.75mg&6 months&Placebo&18 months&6/65&3/21&To avoid error in unit-of-analysis, we calculated the placebo group subjects and recurrences by multiplying the proportion of subjects in the GnRH group out of the total of subjects in all the intervention groups. Truncation was also applied.  Best scenario was assummed to obtain the ITT statistics.\tabularnewline
Huang et al. (2018)\cite{Huangetal2018}&China&2 arms&01/11-12/14&reclassify ASRM stages into two&\mbox{I: 36.41±5.19;} \mbox{C: 36.81±6.92}&laparoscopy&echography&GnRH agonist SC 3.75mg&6 months&No treatment&12 months&6/50&15/50&ITT statistics were reported.\tabularnewline
Yang et al. (2019)\cite{Yangetal2019}&China&2 arms&01/15-03/16&stages III-IV&24-35&laparoscopy&clinical and biochemical assessment&Triptorelin 3.75mg IM&6 months&Expectant&24 months&1/65&9/65&ITT statistics were reported.\tabularnewline
\hline
\end{longtable}}
  \scalebox{1.02}{
  \rotatebox{90}{
  \begin{minipage}[H]{\ht\tablebox}
      \setcounter{table}{0}
      \captionof{table}{Characteristics of Included Studies}
      \captionsetup{skip=12pt}
      \label{tab:tabOne} 
      \unvbox\tablebox
  \end{minipage}
  }
  }
\end{center}
\clearpage

Both analyses (Figures 2(a) and 3) showed negligible heterogeneity, with standard metrics indicating $I^2 = 0.00\%$ and a point estimate of $\tau^2 = 0.00$. $I^2$ expresses the proportion of variability in a meta-analysis which is explained by between-trial heterogeneity rather than by sampling error, and $\tau^2$ the between-trial heterogeneity.\cite{Higginsetal2019}  The complete picture was that the 95\% CI for $\tau^2$ varied widely \mbox{(0.00 to 3)}, revealing that the true level of heterogeneity could be highly uncertain. The choice of the random-effects model was, therefore, appropriate to account for this potential true heterogeneity, highlighting the limitations of relying on point estimates alone when assessing heterogeneity.

\vspace*{-0pt}
\begin{figure}[H]
  \caption{The Primary Model and Its Leave-One-Out Results}
\begin{minipage}[t]{.99\textwidth}
    \centering
    \includegraphics[scale=0.59,trim = 0mm 0mm 0mm 0mm,clip]{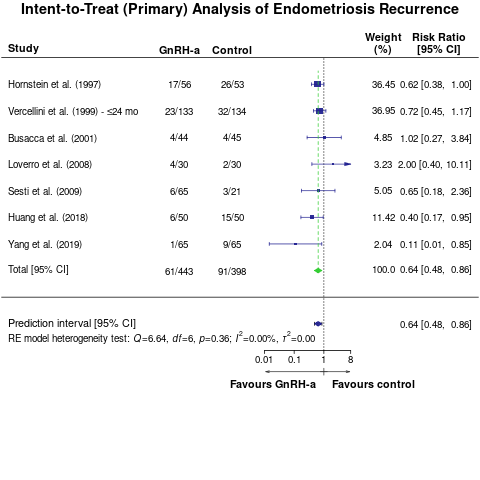}\vspace*{-86pt}
   \flushright (a)
   \flushleft
    \centering
   \includegraphics[scale=0.59,trim = 5mm 10mm 110mm 10mm,clip]{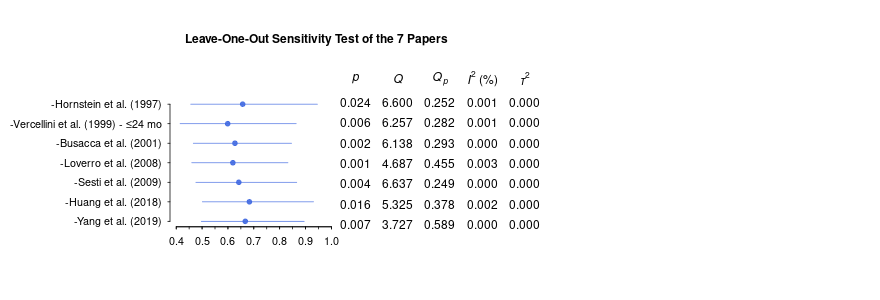}\vspace*{-36pt}
   \flushright (b)
\end{minipage}%
\end{figure}\vspace*{-10pt}

The As-Treated (AT) analysis yielded a pooled effect smaller than the ITT analysis, and the 95\% CI was just 0.01 narrower ({\it cf.} \mbox{Figures~2(a) and 4}). Although the difference was only at the second decimal place, it suggests that the AT analysis slightly overestimates the protective effect of this subclass of GnRH-a agents. At individual study level, Nafarelin nasal 400µg for 6 months was shown to be protective against recurrence after endometriosis surgery in the AT analysis.\cite{Hornsteinetal1997}

\begin{figure}[H]
\begin{minipage}[t]{.99\textwidth}
    \centering
  \caption{Sensitivity Analysis of the 7 papers}
  \includegraphics[scale=0.59,trim = 0mm 10mm 0mm 0mm,clip]{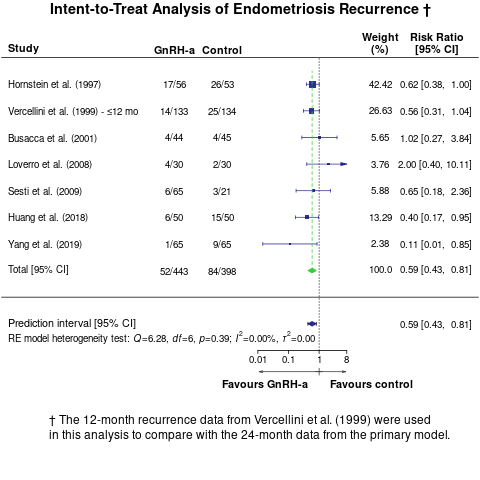}\vspace*{-24pt}
\end{minipage}%
\end{figure}

\begin{figure}[H]
\begin{minipage}[t]{.99\textwidth}
    \centering
  \caption{As-treated Analysis}
  \centering
   \includegraphics[scale=0.59,trim = 0mm 10mm 0mm 0mm,clip]{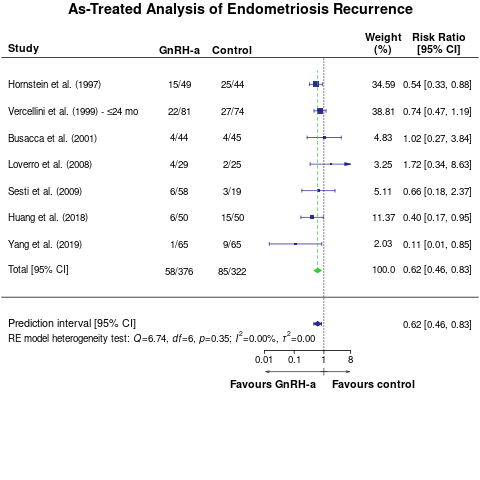}\vspace*{-50pt}
\end{minipage}%
\end{figure}\vspace{-0pt}

The non-parametric bootstrap analysis failed to converge in only 22 of 10,000 replications. It provided a pooled effect estimate, with a 95\% BCa CI of $(-0.84$ to $0.24)$ on the log-scale (or $(0.43$ to $0.79)$ on the RR scale). The close agreement between the standard and bootstrap CIs indicates that the pooled effect is robust and not an artifact of the model's assumptions. The point estimate for $\tau^2$ was $0.00$, with a 95\% CI of $(0.00$ to $3)$. The non-parametric bootstrap for $\tau^2$ could not provide a reliable CI due to the large number of zero-valued replicates. However, the likelihood function of $\tau^2$ reinforced that the between-study heterogeneity was low, with $0.00$ as the peak of the function (Figure 5). The profiling likelihood 95\% CI for $\tau^2$ \mbox{(0.00 to 0.59)} signified uncertainty in heterogeneity, although its upper limit was substantially lower than the CI estimated by the standard meta-analysis. A leave-one-out sensitivity analysis (Figure~2b) further confirmed that \drug\ had a protective effect against \endo\ recurrence after surgery; that is, the pooled effect remained statistically significant and was not disproportionately influenced by the removal of any single trial.

\vspace*{-0pt}
\begin{figure}[H]
  \caption{Between-Study Heterogeneity}
\begin{minipage}[t]{.99\textwidth}
    \centering
    \includegraphics[scale=0.46,trim = 0mm 6mm 0mm 5mm,clip]{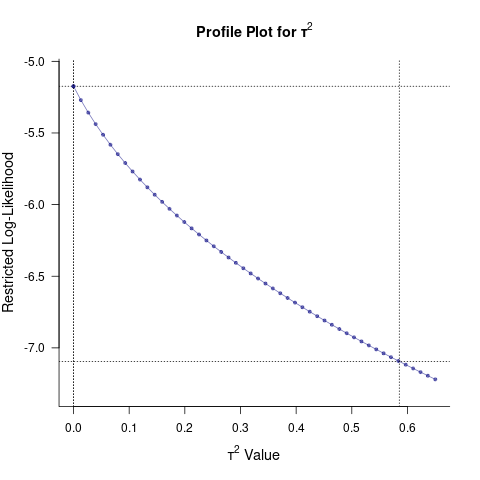}\vspace*{-86pt}
\end{minipage}
\end{figure}\vspace*{-10pt}

A sensitivity analysis was conducted on a 5-paper model that excluded two trials with severe methodological flaws (see \mbox{Figure~6(a)}).\cite{Huangetal2018, Sestietal2009} This model obtained a pooled RR of 0.69 (95\% CI 0.50, 0.94), which was slightly larger and had a wider 95\% CI in comparison with the primary model. Although the point estimate of $\tau^2$ remained 0.00, its 95\% CI \mbox{(0.00, 8.57)} was much wider than that of the primary model, whose upper limit was 3. As \mbox{Figure~6(b)} illustrates, a leave-one-out analysis showed that excluding the trial by \citeauthor{Hornsteinetal1997}\cite{Hornsteinetal1997} from this 5-paper model rendered the pooled effect non-significant ($p = 0.168$). The trial's substantial difference from the pooled effect, combined with the loss of statistical significance, qualifies it as an influential study or a potential outlier.

The risk-of-bias (RoB) summary plot is presented in \mbox{Figure~S.5(a)}. The majority of the trials were rated as having a high RoB in the domains of deviations from intended interventions (D2) and outcome measurement (D4). This suggests potential issues with intervention consistency and outcome assessment across trials. Conversely, low RoB was most prevalent in the domain of missing data (D3) and that of the selection of reported results (D5). No trial scored a high risk in the randomisation process (D1). Individual study scores are shown in \mbox{Figure~S.5(b)}, where all trials were assessed as having a high overall RoB. We assessed eight papers in total. One trial was eligible, but its results were not reliable; therefore, only its RoB could be assessed. Because of the high overall RoB and serious indirectness, the certainty of evidence was downgraded twice in the GRADE assessment (see \mbox{Table S.3}).

\begin{figure}[H]
\begin{minipage}[t]{.99\textwidth}
    \centering
  \caption{A 5-paper Model}
  \includegraphics[scale=0.59,trim = 0mm 28mm 0mm 10mm,clip]{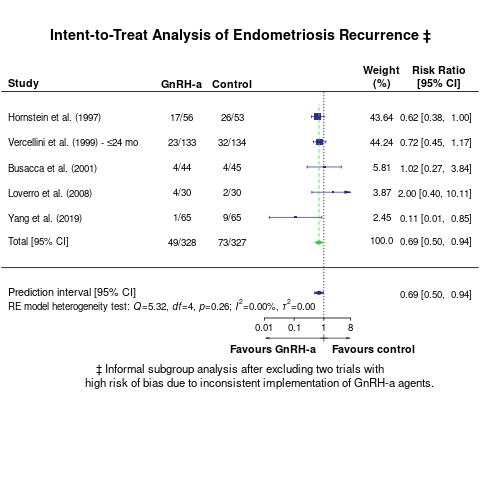}\vspace*{-36pt}
   \flushright (a)
   \flushleft
   \centering
   \includegraphics[scale=0.59,trim = 5mm 10mm 110mm 5mm,clip]{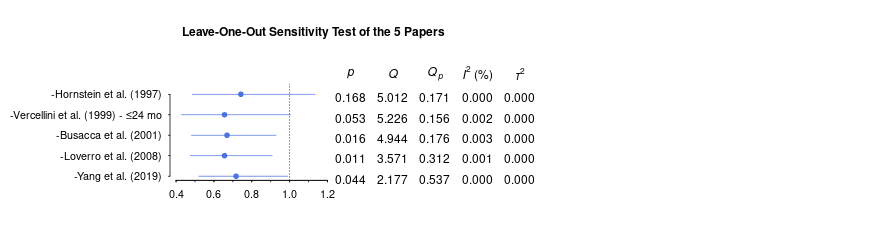}\vspace*{-38pt}
   \flushright (b)
\end{minipage}%
\end{figure}\vspace*{-12pt}

Funnel plot was not performed for this study due to the limited number of included studies ($n<10$ in the sensitivity analysis).\cite{Higginsetal2019} Instead, a Doi plot and the LFK index were calculated to assess for asymmetry.\cite{FuruyaKanamorietal2018} The LFK index was $-0.113$, which falls within the accepted interval of $[-1, 1]$, indicating a lack of statistical evidence for asymmetry (or publication bias). The Doi plot for the seven papers is presented in \mbox{Figure~S.6}, which visually suggested that two trials, and to a lesser extent a third, appeared as potential outliers.

\sec{Discussion}

This systematic review and meta-analysis (SRMA) of seven RCTs identified a promising direction to address the long-standing clinical and methodological question concerning the impact of hormonal therapy on reducing post-operative \endo~recurrence. This study demonstrates that a specific subclass of GnRH-a agents offers a statistically significant protective effect in reducing post-operative recurrence, a finding that directly addresses prevailing clinical scepticism regarding the use of hormonal therapy for managing endometriosis after surgery. Recognising automated tools for SRMA are increasingly free accessible, this paper details a Computational-Assisted Systematic Review and Meta-analysis (CASMA) framework, making it one of the first papers to open the discussion on the use of computer technologies to achieve validated, efficient, reproducible, and transparent evidence synthesis without violating the PRISMA guidelines.

The findings from the seven trials (including 841 patients and 152 inferred recurrences) were subjected to comprehensive validity checks. External validity was established by cross-checking the extracted data against data reported in published reviews -- including a Cochrane Systematic Review, which is widely considered a gold standard in evidence synthesis. While not a gold standard practice, this approach offers a pragmatic and methodologically robust alternative under the circumstances.  The consistency of the results with those reviews implies the extraction was accurate. For internal validity, the primary analysis yielded a pooled RR of 0.64 (a 36\% reduction in risk), with a 95\% CI \mbox{(0.48 to 0.86)}. The close agreement with the bootstrap 95\% CI \mbox{(0.43 to 0.79)} indicates the result is stable and robust. The likelihood profiling method confirmed $\tau^2 = 0.00$ to be a valid point estimate and provided a much narrower 95\% CI than the standard meta-analysis, clarifying the between-study heterogeneity.  Subgroup analyses were performed by replacing a data point in the primary model with one where the follow-up period was shorter, and also on the five papers that had no serious methodological limitations. The leave-one-out analysis on this 5-paper model identified an influential paper, yet the same analysis showed that the primary model remained stable. All analyses, including the As-Treated analysis, consistently suggested that \drug~had a protective effect, and the homogeneity assumption failed to be rejected. These diverse statistical methods were applied to establish internal validity and the robustness of the findings, not for the purposes of data dredging. Publication bias was assessed by the Doi plot with the LFK index rather than a funnel plot to obtain a more reliable measure where the sample size was small.

By focusing on a specific subclass of \drug, the findings provide a focused complement to existing reviews with a broader scope. For example, one review reported that hormonal therapy reduced endometriosis recurrence at both 12 months or less (RR=0.30, 95\% CI \mbox{(0.17 to 0.54)}) and at 13-24 months (RR=0.40, 95\% CI \mbox{(0.27 to 0.58)}), but these findings were coupled with substantial heterogeneity ($I^2=58\%$ and $57\%$ respectively).\cite{Chenetal2020} Another review found a protective effect across a broader range of hormonal therapies (RR=0.44, 95\% CI \mbox{(0.30 to 0.66)}), though with low reported heterogeneity ($I^2=3\%$).\cite{Zakharietal2021} A direct comparison of the extent of this reduction is challenging, as the estimations are based on different methodologies and the definition of recurrence varies widely across studies.\cite{Ceccaronietal2019} A previous review had a similar scope to this paper,\cite{Zhengetal2016} but the methodology was prone to unit-of-analysis errors and included studies that deviated from standardised definitions. By applying control-group splitting and excluding trials with potential confounding factors, the present analysis provides a more internally consistent assessment of this subclass of \drug.

This study updates the evidence on endometriosis recurrence through the applications of recently standardised definitions\cite{Tomassettieta2021}, current methodological standards and well-established computational techniques. The CASMA framework adds a process of pre-screening prior to manual paper selection. It only involves applying a semi-automated text-matching approach -- fuzzy matching and regex -- to efficiently exclude 812 irrelevant records. From fetching 33\,444 records to completing the pre-screening, it took less than 11 days to obtain a manageable collection of 29 records. Inspired by a previously proposed semi-automated method for analysing unstructured medical data,\cite{Tsang2020} this approach balances expert-driven screening with automated solutions and can be applied to larger or more complex datasets without violating PRISMA standards. The efficiency of the CASMA pre-screening minimised the subsequent manual workload. Crucially, this framework eases the decision-making process by avoiding overloading experts with information, allowing them to efficiently adapt and integrate their clinical and methodological expertise via iterative search string refinement. This approach is preferred to applying fully automated but standardised solutions, where researchers are only given the option of accepting or rejecting the proposed evidence.

The certainty of evidence, as assessed with the GRADE framework, was rated as ``Low'' \mbox{(see Table S.3)}. This was primarily due to a very serious RoB and serious indirectness. Bias arose most prominently from deviations from intended interventions and outcome measure, while indirectness reflected variations in recurrence definitions, limiting the generalisability of the findings. The small sample size prevented both a formal subgroup analysis and a full Qualitative Comparative Analysis (QCA). Although QCA was a compelling tool to explore necessary and sufficient conditions for recurrence, the limited number of studies ($N=7$) and high data sparsity precluded a meaningful interpretation. This paper demonstrates a case of applying CASMA to synthesise robust evidence from a large body of literature with much ambiguity circulating, offering a promising framework to synthesise the current and best evidence to facilitate Evidence-Based Medicine practice.

\sec{Competing Interests}
The author declares no competing interests.

\sec{Funding Statement}
This research received no specific grant from any funding agency in the public, commercial, or not-for-profit sectors.

\sec{Acknowledgements}
I am deeply indebted to Professor Bruno Falissard of Université Paris-Saclay for his intellectual guidance on this project and throughout the master degree program. My thanks also extend to the dedication of other professors and Mr. Fares Youbi for his generous assistance. I would like to thank Miss Meriem Souici for her assistance with paper selection and some data extraction. I sincerely thank Professor Richard F. Heller of Universities of Manchester, UK, and Newcastle, Australia for recruiting me as a volunteer tutor for his charity foundation; the experience inspired me to find my knowledge and skills for health-related research.

The opinions expressed in this article are those of the author and do not necessarily reflect the views or policies of the affiliated institutions. Authorship is defined by the ICMJE criteria.

\sec{Ethical Approval}
This systematic review and meta-analysis was completed as part of the requirements for a Master of Public Health (MPH) degree in Methodology and Biostatistics for Biomedical Research, with research directions approved by Professor Bruno Falissard at the Faculty of Medicine, Université Paris-Saclay. Since this study did not involve any human or animal subjects or identifiable data, it was exempt from further institutional review.

\sec{Protocol Registration and Data/Code Availability}
The protocol for this research is registered on OSF. The DOI is \url{https://doi.org/10.17605/OSF.IO/R2DFA}. The timestamp is 5 September 2025.

All data used were extracted from published studies. The data needed to reproduce the results are included in the main content. Other data and the {\tt R} codes will be deposited in the public OSF repository when this manuscript is published in a peer-reviewed journal.

\bibliography{Devoir/06-citations}
\bibliographystyle{elsarticle-nbr-jecp}


\setcounter{figure}{0}
\renewcommand\thefigure{S.\arabic{figure}}
\setcounter{table}{0}
\renewcommand\thetable{S.\arabic{table}}
\newpage
\sec{Supplementary Material}

\subsec{A Research Domain Grows Exponentially}
\begin{minipage}[H]{.52\textwidth}
 \leftskip  -24pt
 \parbox{1\textwidth}{\captionof{figure}{The Growth of the Publications in Endometriosis}}\vspace{-0pt}
 \includegraphics[scale=0.49,trim = 0mm 0mm 0mm 5mm,clip]
{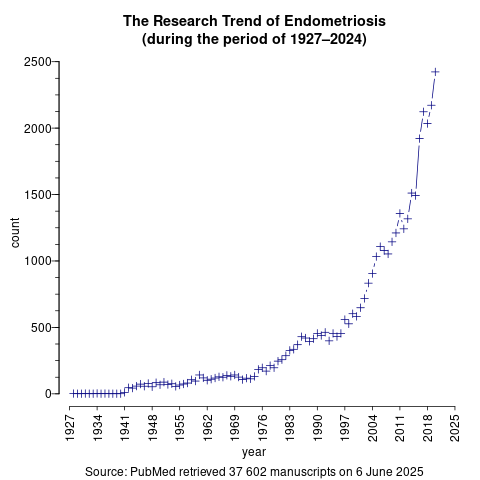}
\end{minipage}
\begin{minipage}[H]{.48\textwidth}
 \parbox{1.3\textwidth}{\captionof{figure}{An Approximation of Endometriosis Research in the Pipeline$\dagger$}}\vspace{-0pt}
 \includegraphics[scale=0.49,trim = 0mm 0mm 0mm 5mm,clip]
{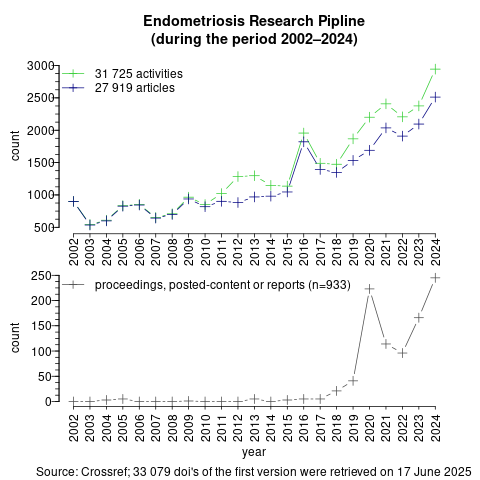}
\end{minipage}\vspace*{6pt}
\hspace*{.55\textwidth}
\parbox{.56\textwidth}{\fontsize{10}{9}\selectfont
$\dagger$ Due to limited computational resources, the records could only be deduplicated using the prefixes of the DOIs. This method is not as accurate as fuzzy matching.
}

\newpage
\subsec{Extracting Records and Selection Process}
Search terms were mutually agreed upon by ST and MS. The primary literature search (for meta-analyses of RCTs) was conducted from \mbox{6 June} to 14 July 2025. Records were collected from multiple sources: PubMed yielded 37\,602 documents on \mbox{6 June 2025}; Google Scholar (GS) retrieved 583\,000 documents on 10 June 2025; Crossref retrieved 33\,169 documents on 17 June 2025; and Scopus retrieved 52\,561 documents on \mbox{17 June 2025}.

Based on repeated attempts, it was determined by ST that the following syntax was the most reliable for screening out relevant records from PubMed:

\noindent ``((((endometriosis) AND ((((((((((((((Hormonal contraceptive) OR (Progestin therapy)) OR (Gonadotropin-releasing hormone)) OR (Aromatase inhibitor)) OR (add-back therapy)) OR (non-steroidal anti-inflammatory drug)) OR (steroid)) OR (androgen))) OR (hormone replacement)) OR (hormone therapy)) OR (hormone-related therapy)) OR (hormone suppression)) OR (hormonal alteration))) AND ((randomised controlled trial) OR (clinical trial))) AND (meta analysis))''

Subsequently, ST adapted this keyword set to search Google Scholar (GS). GS retrieved over 4\,160 records with these keywords. Given that its \texttt{robots.txt} file restricts web scraping, a maximum of 100 records could be obtained. To heighten the chance of retrieving relevant manuscripts within that limit, ST included outcome keywords in the search. The record count dropped from 4\,160 to 1\,470. ST fetched 140 webpages and prioritised the extraction of the most relevant records. The applied keyword set was as follows:

\noindent ``((((endometriosis) AND ((((((((((((((Hormonal contraceptive) OR (Progestin therapy)) OR (Gonadotropin-releasing hormone)) OR (Aromatase inhibitor)) OR (add-back therapy)) OR (non-steroidal anti-inflammatory drug)) OR (steroid)) OR (androgen))) OR (hormone replacement)) OR (hormone therapy)) OR (hormone-related therapy)) OR (hormone suppression)) OR (hormonal alteration))) AND ((randomised controlled trial) OR (clinical trial))) AND (meta-analysis)) AND ((cancer) OR (recurrence) OR (adverse effect))''

\noindent Data from the 140 records were extracted using an {\tt R} programme written by ST. A dissimilarity matrix of all the titles was computed using fuzzy matching. Levenshtein distance was the computational basis for this process. It yielded results similar to those obtained by the native \texttt{agrep()} {\tt R} function. Any count smaller than 5, but not on the diagonal of the matrix, was considered a potentially duplicated pair. The threshold was set to 5 because, on average, an English word contains 5 letters. The results suggested the elimination of 25 records. This approach was applied to check for duplicates both within and across databases. A manual review of the records suggested that the relevance of subsequent entries was low, indicating a low chance of having missed crucial records.

Crossref only allowed case-insensitive searches with the exact word, ``endometriosis.'' When more terms were added, many more titles were retrieved than were relevant to the term ``endometriosis.'' The system did not allow fetching the titles and abstracts of records with full-text. ST first fetched the title, creation date, DOI, publisher, type, and indexed information for each of 33\,169 records. ST then fetched titles and DOIs for the records with full text. Of the $27\,886$ records with full-text, $565$ fell into the categories of posted-content, proceedings, and reports (referred to as grey literature). $42$ duplicated records were eliminated. Regex searches were performed on $388$ abstracts and their respective titles. For titles without abstracts, ST reviewed the abstracts of the records indexed by the chosen syntaxes. A variation of this approach was applied to the records obtained from other databases. \mbox{Figure~S.2} shows the research trend after records with unique DOI prefixes. This only gave an approximate view about the trend, because a single record could be registered on different platforms. A more accurate view could not be obtained due to insufficient computational power to apply fuzzy matching to the titles.

Only six meta-analyses that fitted the research topic were identified. Of these, only three reviewed RCT primary papers where the interventions were Diegogest, a GnRH-a, and a \mbox{GnRH-a} with Chinese medicine. The latter was not considered a potential topic because Chinese medicine is a broad field, which effectively tested an unknown agent. Due to a tight timeline, the decision was made to hand-search the RCTs within meta-analyses from a related domain and perform a free-text search. The eligibility was confirmed during data extraction rather than independent review.

\newgeometry{left=2.5cm, right=2.5cm, top=1.5cm, bottom=2cm}
\subsec{Manual Process of Paper Selection}
{  \centering
  \captionsetup{skip=12pt}
  \captionof{table}{The 29 Records Shortlisted Following the Computational-Assisted Process}
\newcolumntype{P}[1]{>{\raggedright\arraybackslash}p{#1}}
\fontsize{7.5}{7}\selectfont
\renewcommand*{\arraystretch}{1.2}
\setlength{\tabcolsep}{1.5pt}
\begin{longtable}{P{1.5cm}P{1.5cm}P{3.8cm}*{6}{c}P{3.5cm}P{2.42cm}}
\hline
database & author & title & \multicolumn{2}{c}{round 1} & assess & \multicolumn{2}{c}{round 2} & potential & action & note \\
& & & \multicolumn{1}{c}{MS}& \multicolumn{1}{c}{ST}& \multicolumn{1}{c}{}& \multicolumn{1}{c}{MS}& \multicolumn{1}{c}{ST}& & & \\
\hline
\endhead

  \hline
  \multicolumn{11}{l}{\footnotesize Continued on the next page}\\
  \endfoot
  \endlastfoot

  crossref & Shen et al. (2020) & Fertility outcomes of deep infiltrating endometriosis with fertility desire—a meta analysis & 0 & -1 & 0 & -1 & -1 & -1 & Studied fertility desire only -- Outside research scope & excluded \\
  crossref & Qing et al. (2004) & Systematic Review and Meta-analysis on the Effect of Adjuvant Gonadotropin-releasing Hormone Agonist (GnRH-a) on Pregnancy Outcomes in Women with Endometriosis Following Conservative Surgery. & -1 & -1 & -1 & -1 & -1 & -1 & Studied fertility desire only -- outside research scope & excluded at round 1 \\
  GS & Yang (2024) & Endometriosis and aspirin: a systematic review & -1 & -1 & -1 & -1 & -1 & -1 & The study compared non-human samples. & excluded at round 1 \\
  GS & Mikuš et al. (2022) & State of the art, new treatment strategies, and emerging drugs for non-hormonal treatment of endometriosis: a systematic review of randomized control trials & -1 & 1 & -1 & -1 & -1 & -1 & Three subgroups: antiangiogenic agents, immunomodulators, and natural components. Pelvic pain is not a chosen research topic. & excluded at round 1 \\
  PubMed & Johnstone et al. (2015) & Controversies in the Management of Endometrioma: To Cure Sometimes, to Treat Often, to Comfort Always? & -1 & -1 & -1 & -1 & -1 & -1 & Its abstract is unstructured. The MeSH terms indicated that it is a meta-analysis. The abstract suggests that recurrence was not an outcome measure. & excluded at round 1 \\
  PubMed & Chen et al. (2020) & Effect of melatonin for the management of endometriosis: A protocol of systematic review and meta-analysis. & 0 & 0 & 0 & -1 & -1 & -1 & The study focuses on the effect of melatonin; it is unclear if both beneficial and adverse effects were examined. & excluded \\
  PubMed & Deng et al. (2020) & Chinese herbal medicine for previous cesarean scar defect: A protocol for systematic review and meta-analysis. & -1 & -1 & -1 & -1 & -1 & -1 & The study focuses on Chinese herbal medicine for previous Caesarean scar defect, not endometriosis. & excluded at round 1 \\
  PubMed & Zhang et al. (2021) & The efficacy and safety of Kuntai capsule combined with leuprorelin acetate in the treatment of endometriosis: A protocol for systematic review and meta-analysis. & 1 & 1 & 1 & -1 & -1 & -1 & Multiple, heterogeneous interventions. Translating the evidence into clinical practice would be difficult.. & excluded \\
  PubMed & Gao et al. (2022) & Salvia miltiorrhiza-Containing Chinese Herbal Medicine Combined With GnRH Agonist for Postoperative Treatment of Endometriosis: A Systematic Review and meta-Analysis. & 1 & 1 & 1 & -1 & 1 & 1 & Chinese Herbal Medicine is a broad area. It is akin to studying the relationship between an unspecified agent of unspecified quantity and endometriosis recurrence. & excluded \\
  PubMed & Allahqoli et al. (2024) & Neuropelveology for Endometriosis Management: A Systematic Review and Multilevel Meta-Analysis. & -1 & -1 & -1 & -1 & -1 & -1 & The focus is on a surgical method (Neuropelveology) & excluded at round 1 \\
  PubMed & Piacenti et al. (2025) & Dienogest vs. combined oral contraceptive: A systematic review and meta-analysis of efficacy and side effects to inform evidence-based guidelines. & 1 & 1 & 1 & -1 & -1 & -1 & The author analysed RCT and observational studies together. & excluded \\
  Scopus & Lata and Sarwar (2014) & Effectiveness of conservative surgery and adjunctive hormone suppression therapy versus surgery alone in the treatment of symptomatic endometriosis: A systematic review with meta-analysis & -1 & 1 & -1 & -1 & -1 & -1 & The study compares adjunctive hormone suppression therapy to surgery alone, measuring pelvic pain and recurrence. & excluded at round 1 \\
  Scopus & Chen (2014) & Effectiveness and safety of postoperative GnRH-a versus laparoscopy alone for endometriosis: A meta-analysis & 0 & 0 & 0 & -1 & -1 & -1 & Full-text paper was unavailable; excluded due to an inability to contact the authors. & excluded \\
  Scopus & Liu (2021) & Dienogest as a Maintenance Treatment for Endometriosis Following Surgery: A Systematic Review and Meta-Analysis & 1 & 1 & 1 & -1 & -1 & -1 & Comparators: Levonorgestrel-releasing intrauterine system (LNG-IUS) and gonadotropin-releasing hormone analogs (GnRH-a), or non-treatment (NT). Recurrence is an outcome. & excluded \\
  Scopus & Whelan (2022) & Risk Factors for Ovarian Cancer: An Umbrella Review of the Literature & -1 & -1 & -1 & -1 & -1 & -1 & It is a meta-analysis of cohort studies, which is not the study design of interest. & excluded at round 1 \\
  Scopus & Ivanov (2023) & The issues of endometriosis hormonal treatment in reproductive age women & -1 & -1 & -1 & -1 & -1 & -1 & The full text is written in Russian & excluded at round 1 \\
  Scopus & de Souza Gaio et al. (2025) & Clinical effectiveness of progestogens compared to combined oral contraceptive pills in the treatment of endometriosis: A systematic review and meta-analysis & 1 & 1 & 1 & 1 & 1 & 1 & All studies are RCTs. The study compares Progestogens vs OC, and measures pelvic pain, dysmenorrhea, and psychological symptoms. & independent topic \\
  Scopus & Li (2024) & Efficacy and safety of dienogest in the treatment of endometriosis: a meta-analysis & -1 & -1 & -1 & -1 & -1 & -1 & The full text is written & excluded at round 1 \\
  Scopus & Thiel (2024) & The Effect of Hormonal Treatment on Ovarian Endometriomas: A Systematic Review and Meta-Analysis & -1 & -1 & -1 & -1 & -1 & -1 & Mixed study types & excluded at round 1 \\
  Scopus & Shi (2022) & Effect and safety of drospirenone and ethinylestradiol tablets (II) for dysmenorrhea: A systematic review and meta-analysis & -1 & -1 & -1 & -1 & -1 & -1 & Dysmenorrhea is not endometriosis & excluded at round 1 \\
  Scopus & Peng (2021) & Dydrogesterone in the treatment of endometriosis: evidence mapping and meta-analysis & -1 & -1 & -1 & -1 & -1 & -1 & Mixed study types & excluded at round 1 \\
  Scopus & Chen (2020) & Pre- and postsurgical medical therapy for endometriosis surgery & -1 & -1 & -1 & -1 & -1 & -1 & The study states hormonal suppression is effective but most studies are covered by other review papers. Hormonal suppression is a broad topic. It does not suit a project with a tight timeline. & excluded at round 1 \\
  Scopus & Zakhari (2020) & Dienogest and the Risk of Endometriosis Recurrence Following Surgery: A Systematic Review and Meta-analysis & 1 & 1 & 1 & -1 & -1 & -1 & The paper compares dienogest to expectant management, but analyses observational studies. & excluded \\
  Scopus & Jeng (2014) & A comparison of progestogens or oral contraceptives and gonadotropin-releasing hormone agonists for the treatment of endometriosis: A systematic review & -1 & -1 & -1 & -1 & -1 & -1 & The author noted that proceeding to a meta-analysis was impossible. & excluded at round 1 \\
  Scopus & Wu et al. (2014) & Clinical efficacy of add-back therapy in treatment of endometriosis: A meta-analysis & -1 & 1 & 0 & -1 & 1 & 1 & The study compares GnRH-a with add-back therapy versus GnRH-a alone and addresses side effects such as osteoporosis and menopausal syndrome. & may match Gao et al. \\
  Scopus & Wu, Wu and Liu (2013) & Oral contraceptive pills for endometriosis after conservative surgery: A systematic review and meta-analysis & -1 & 1 & 0 & -1 & 1 & 1 & The study compares OC to no OC and other drugs including gestrinone, mifepristone, or GnRH-a, measuring recurrence and remission. & studied recurrence, but can't match Zheng et al. (2016) \\
  Scopus & Wong and Lim (2011) & Hormonal treatment for endometriosis associated pelvic pain & 1 & 1 & 1 & -1 & 1 & 1 & It is a review of RCTs that synthesizes evidence from three trial groups. The trials that compared combined oral contraceptives with progestogen might align with another meta-analysis. However, the outcome measures were dysmenorrhea, not endometriosis recurrence. & excluded \\
  Keyword search & Zheng et al. (2016) & Can postoperative GnRH agonist treatment prevent endometriosis recurrence? A meta-analysis? & 1 & 1 & 1 & -1 & 1 & 1 & It reviewed RCTs and recurrence of managing operated endometriosis patient with GnRH agonist. & independent topic \\
  Keyword search & Zakhari et al. (2021) & Endometriosis recurrence following post-operative hormonal suppression: a systematic review and meta-analysis & 1 & 1 & 1 & 1 & -1 & -1 & The paper compares hormonal therapies with expectant management, and recurrence is an outcome. It analyses both RCTs and observational studies together. & It is of reference value (mixed study types) \\
  \hline
\end{longtable}
  \leftskip -0pt
  \rightskip -18pt
  \parbox{1\textwidth}{\fontsize{10}{10}\selectfont
   N.B. A three-point scoring system was applied for records during the screening process: $-1$ for rejected, 0 for unclear, and 1 for included (full-text retrieval). In both screening stages, records with a score of 0 were re-evaluated.
  }
}
\restoregeometry
\clearpage

\newpage
\subsec{Inter-Rater Reliability Plots}
 \parbox{1\textwidth}{\captionof{figure}{Distribution of Bootstrapped Inter-Rater Reliability at Two Stages}}\vspace{-0pt}
\begin{minipage}[H]{.52\textwidth}
  \leftskip  -24pt
  \vspace*{6pt}
  \includegraphics[scale=0.49,trim = 0mm 10mm 0mm 5mm,clip]{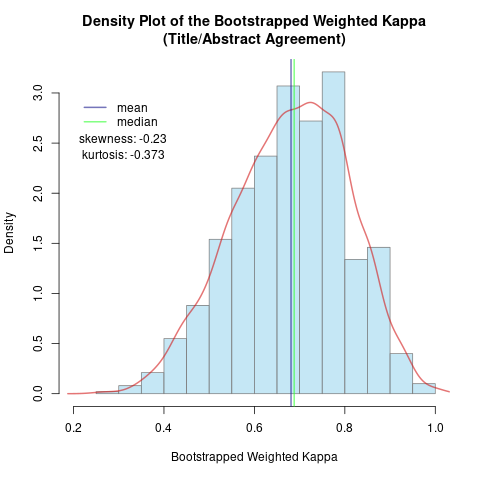}
  \vspace*{-24pt}
  \flushright (a)
\end{minipage}
\begin{minipage}[H]{.48\textwidth}
 \leftskip  10pt
 \includegraphics[scale=0.49,trim = 0mm 10mm 0mm 5mm,clip]{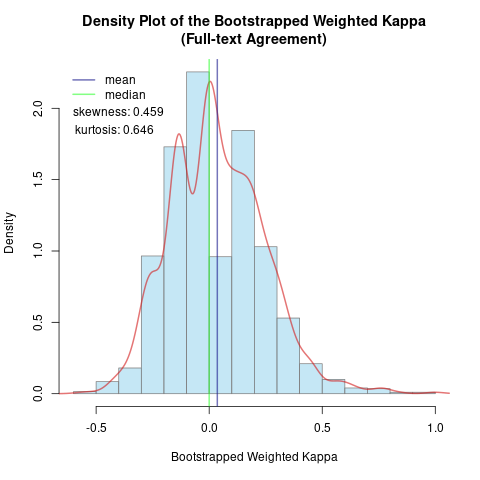}
  \vspace*{-42pt}
  \flushright (b)
\end{minipage}\vspace*{12pt}
The two density graphs show the distribution of the weighted absolute Kappa ($\kappa$) obtained from $2\,000$ non-parametric bootstrap replications. For Title/Abstract screening, the distribution exhibited slight left skew ($-0.23$) and was mesokurtic ($-0.373$). The mean $\kappa$ was $0.68$, and the median $\kappa$ was $0.69$. For Full-text screening, the distribution was right skewed ($0.46$), mesokurtic ($0.65$), and visually exhibited bimodality. The mean $\kappa$ was $0.04$, and the median $\kappa$ was $0.00$. Both distributions are characterized by having skewness in an excellent range ($[-1,+1]$) and kurtosis in a generally acceptable range ($[-2,+2]$).\cite{Hairetal2022} The agreement for the Title/Abstract review was substantial, but that for the Full-text review was poor.\cite{LandisKoch1977}

\newpage
\onehalfspacing
\begin{adjustwidth}{-10pt}{-10pt}
\subsec{\mbox{Deduplication through Fuzzy Matching}}
  \begin{minipage}[H]{.99\textwidth}
      \vspace*{12pt}
      \captionsetup{skip=-6pt}
      \setcounter{table}{1}
      \captionof{table}{An Example of Identifying String Dissimilarities}
    \newcolumntype{P}[1]{>{\raggedright\arraybackslash}p{#1}}
\newcolumntype{C}[1]{>{\centering\arraybackslash}p{#1}}
\newcolumntype{H}{>{\setbox0=\hbox\bgroup}l<{\egroup}@{}}
\renewcommand*{\arraystretch}{1}
\setlength{\tabcolsep}{1.5pt}
\fontsize{8}{7.5}\selectfont
\begin{center}
\begin{longtable}{P{5.5cm}cccccccc}\hline
\noalign{\vspace{3pt}} 
\multicolumn{1}{c}{\rotatebox{90}{}}&%
\multicolumn{1}{c}{\rotatebox{90}{%
 \parbox[t]{5.5cm}{%
   Post-operative GnRH analogue treatment after conservative surgery for symptomatic endometriosis stage III--IV: a randomized controlled trial}}}&%
\multicolumn{1}{c}{\rotatebox{90}{%
 \parbox[t]{5.5cm}{%
   Use of nafarelin versus placebo after reductive laparoscopic surgery for endometriosis.}}}&%
\multicolumn{1}{c}{\rotatebox{90}{%
 \parbox[t]{5.5cm}{%
   Clinical efficacy and safety of gonadotropin-releasing hormone agonist combined with laparoscopic surgery in the treatment of endometriosis}}}&%
\multicolumn{1}{c}{\rotatebox{90}{%
 \parbox[t]{5.5cm}{%
   A randomized study comparing triptorelin or expectant management following conservative laparoscopic surgery for symptomatic stage III--IV endometriosis}}}&%
\multicolumn{1}{c}{\rotatebox{90}{%
 \parbox[t]{5.5cm}{%
   Recurrence rate of endometrioma after laparoscopic cystectomy: a comparative randomized trial between post-operative hormonal suppression treatment or dietary therapy vs. placebo}}}&%
\multicolumn{1}{c}{\rotatebox{90}{%
 \parbox[t]{5.5cm}{%
   A gonadotrophin-releasing hormone agonist compared with expectant management after conservative surgery for symptomatic endometriosis}}}&%
\multicolumn{1}{c}{\rotatebox{90}{%
 \parbox[t]{5.5cm}{%
   A gonadotrophin-releasing hormone agonist compared with expectant management after conservative surgery for symptomatic endometriosis}}}&%
\multicolumn{1}{c}{\rotatebox{90}{%
 \parbox[t]{5.5cm}{%
   Laparoscopic surgery combined with GnRH agonist in endometriosis}}}\tabularnewline
\noalign{\vspace{2pt}} 
\hline
Post-operative GnRH analogue treatment after conservative surgery for symptomatic endometriosis stage III--IV: a randomized controlled trial&0&98&114&105&134&99&99&107\tabularnewline
Use of nafarelin versus placebo after reductive laparoscopic surgery for endometriosis.&98&0&84&93&136&86&86&58\tabularnewline
Clinical efficacy and safety of gonadotropin-releasing hormone agonist combined with laparoscopic surgery in the treatment of endometriosis&114&84&0&92&129&84&84&95\tabularnewline
A randomized study comparing triptorelin or expectant management following conservative laparoscopic surgery for symptomatic stage III--IV endometriosis&105&93&92&0&137&76&76&112\tabularnewline
Recurrence rate of endometrioma after laparoscopic cystectomy: a comparative randomized trial between post-operative hormonal suppression treatment or dietary therapy vs. placebo&134&136&129&137&0&136&136&138\tabularnewline
A gonadotrophin-releasing hormone agonist compared with expectant management after conservative surgery for symptomatic endometriosis&99&86&84&76&136&0&0&90\tabularnewline
A gonadotrophin-releasing hormone agonist compared with expectant management after conservative surgery for symptomatic endometriosis&99&86&84&76&136&0&0&90\tabularnewline
Laparoscopic surgery combined with GnRH agonist in endometriosis&107&58&95&112&138&90&90&0\tabularnewline
\hline
\end{longtable}\end{center}

  \end{minipage}\vspace*{-30pt}
\parbox{1\textwidth}{\fontsize{10}{10}\selectfont 
\mbox{N.B. The 6th} and 7th titles exemplify the results of applying fuzzy matching to identify duplicated titles. Where the number of titles is enormously large, any score below 5 and located off the diagonal of the matrix represents a pair of titles with close similarity and may be further inspected.
}
\end{adjustwidth}

\newpage
\subsec{\mbox{Network of Publications}}
\begin{figure}[H]
  \caption{The Findings of an AI Research Tool\cite{ResearchRabbit2021}}
\begin{minipage}[t]{.99\textwidth}
    \centering
    \includegraphics[scale=0.45,trim = 0mm 0mm 10mm 30mm,clip]{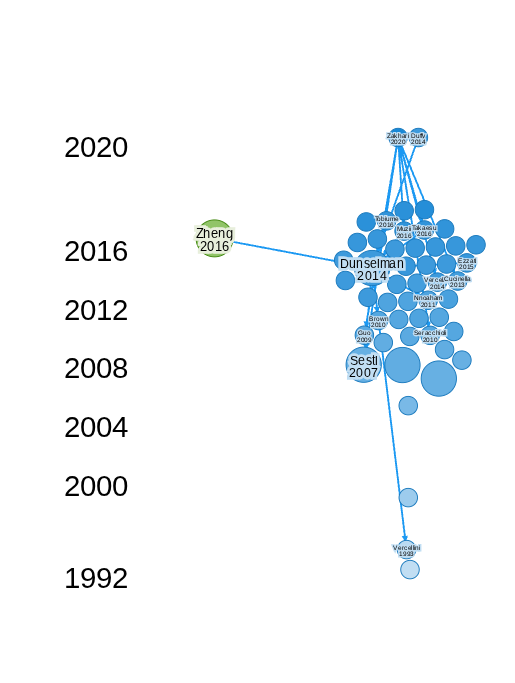}
\end{minipage}%
\end{figure}\vspace{-24pt}
{\fontsize{12}{10}\selectfont N.B. A targeted search was conducted on 14 July 2025 to verify that the meta-analysis by \citeauthor{Zhengetal2016}\cite{Zhengetal2016} was the most recent meta-analysis of RCTs on our chosen topic. The two other meta-analyses published around 2020 included one that synthesised both RCTs and observational studies\cite{Zakharietal2021} and another that focused on surgeries. Consequently, the decision to perform a targeted search rather than running another full selection process was appropriate and not driven solely by the project timeline.
}

\newpage
\subsec{\mbox{Risk of Bias}}
\begin{figure}[H]
\begin{minipage}[t]{.99\textwidth}
    \centering
  \caption{Summary of Risk of Bias (RoB2)}
   \includegraphics[scale=0.63,trim = 0mm 0mm 0mm 0mm,clip]{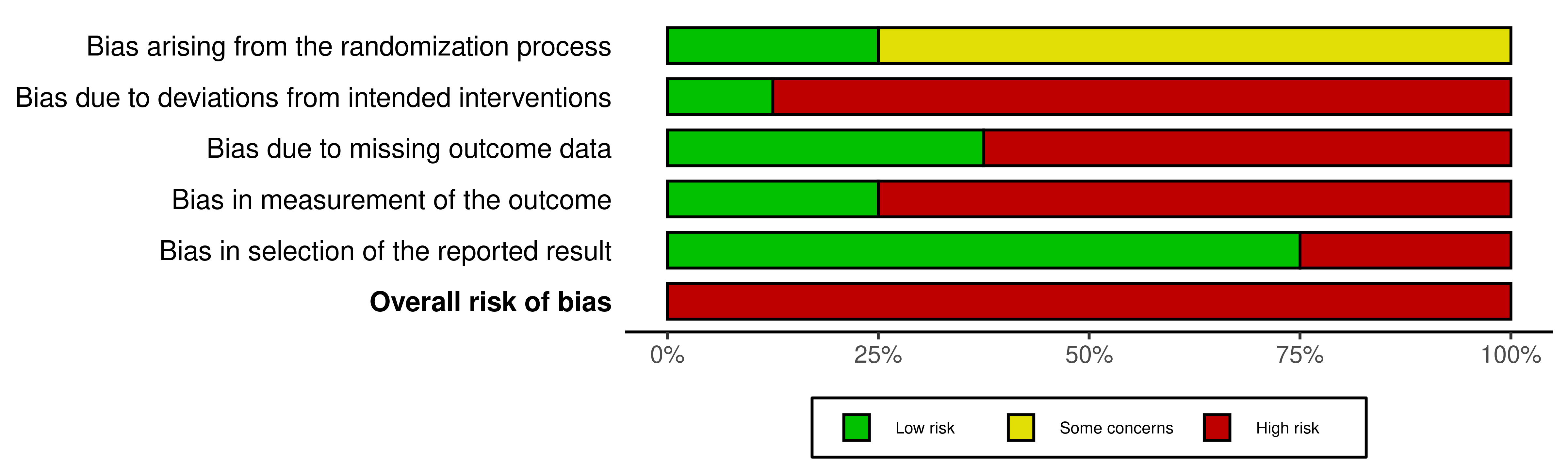}\vspace*{-24pt}
   \flushright (a)\vspace*{-6pt}
    \flushleft
    \centering
    \includegraphics[scale=0.55,trim = 0mm 0mm 0mm 0mm,clip]{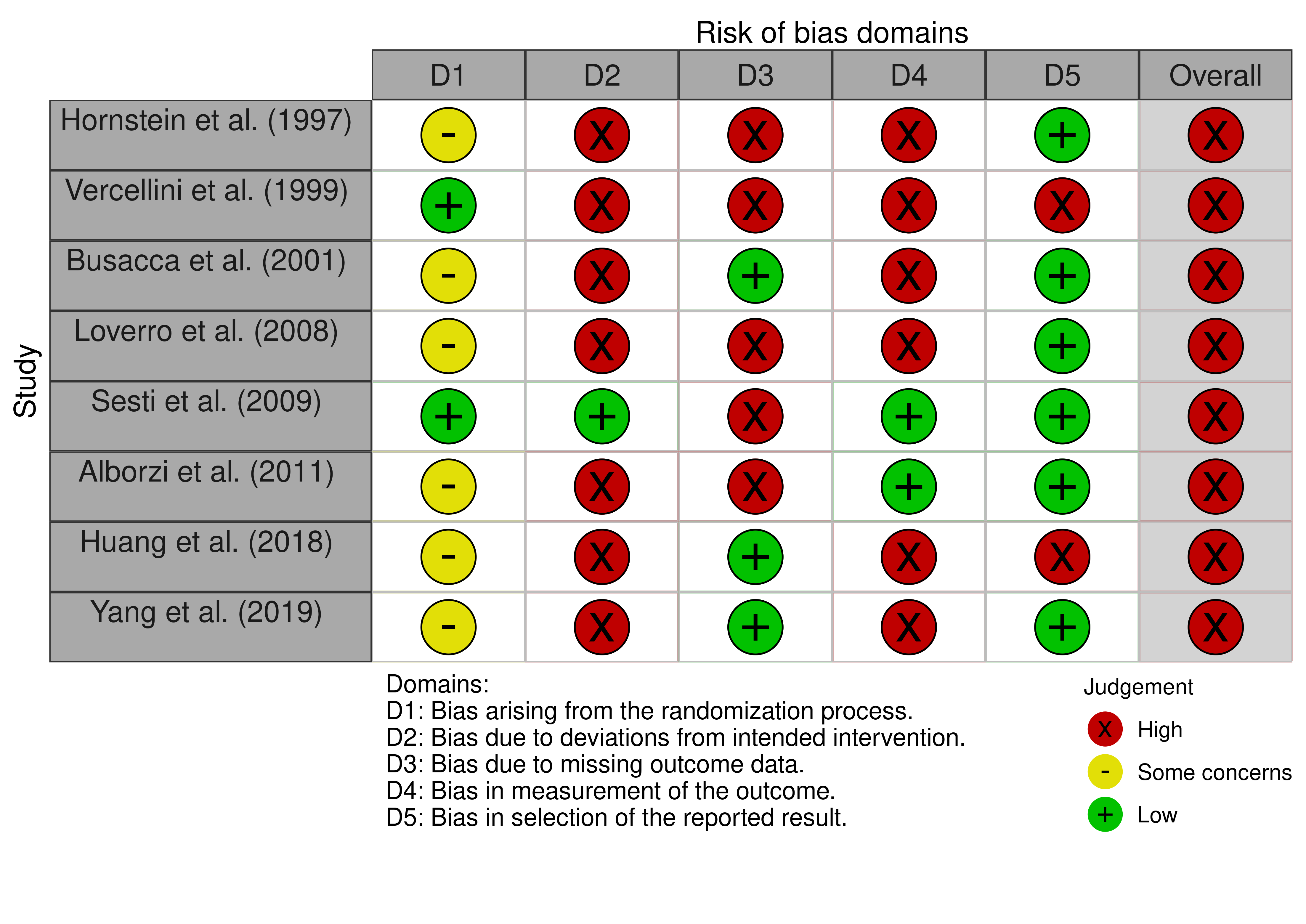}\vspace*{-36pt}
    \flushright (b)
\end{minipage}%
\end{figure}\vspace*{-0pt}

\subsec{\mbox{Publication Bias}}
\begin{figure}[H]
\begin{minipage}[t]{.99\textwidth}
    \centering
  \caption{Less Restrictive Measures of Publication Bias}
    \includegraphics[scale=0.52,trim = 0mm 12mm 0mm 12mm,clip]{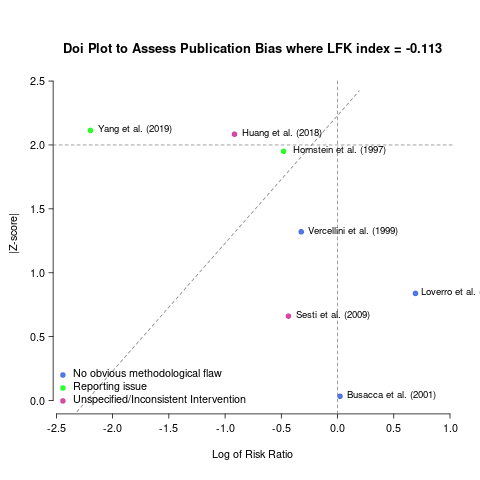}
\end{minipage}%
\end{figure}\vspace*{-5mm}

\newpage
\subsec{Non-parametric Bootstrap Analysis}
The following code was applied to obtain the non-parametric bootstrap results to verify the pooled risk ratio, between-study heterogeneity and the corresponding 95\% CIs.

\begin{lstlisting}
# Non-Parametric
boot.func <- function(dat, indices) {
   sel <- dat[indices,]
   res <- try(suppressWarnings(rma(yi, vi, data=sel)), silent=TRUE)
   if (inherits(res, "try-error")) {
      NA
   } else {
      c(coef(res), vcov(res), res$tau2, res$se.tau2^2)
   }
}
set.seed(print(seed.nbr<-sample(2^16,1))) #14453
(res.boot <- boot::boot(dat, boot.func, parallel = "multicore",R=10000))
range(res.boot$t[,3][!is.na(res.boot$t[,3])])
attr(res.boot,"seed.nbr")<-seed.nbr
saveRDS(res.boot,"MA-Result_bootstrap_np.rds")
\end{lstlisting}

\clearpage
\begin{sideways}
    \begin{minipage}{1\textheight}
        \vspace*{-72pt}
        \subsec{GRADE Assessment}
        \vspace*{12pt}
        \setcounter{table}{2}
       \captionof{table}{Summary of Findings$\dagger$$\ddagger$}
        \label{tab:tabSoF}
        \hspace*{-18pt}
        \includegraphics[scale=0.9, trim = 2mm 45mm 2mm 2mm, clip]{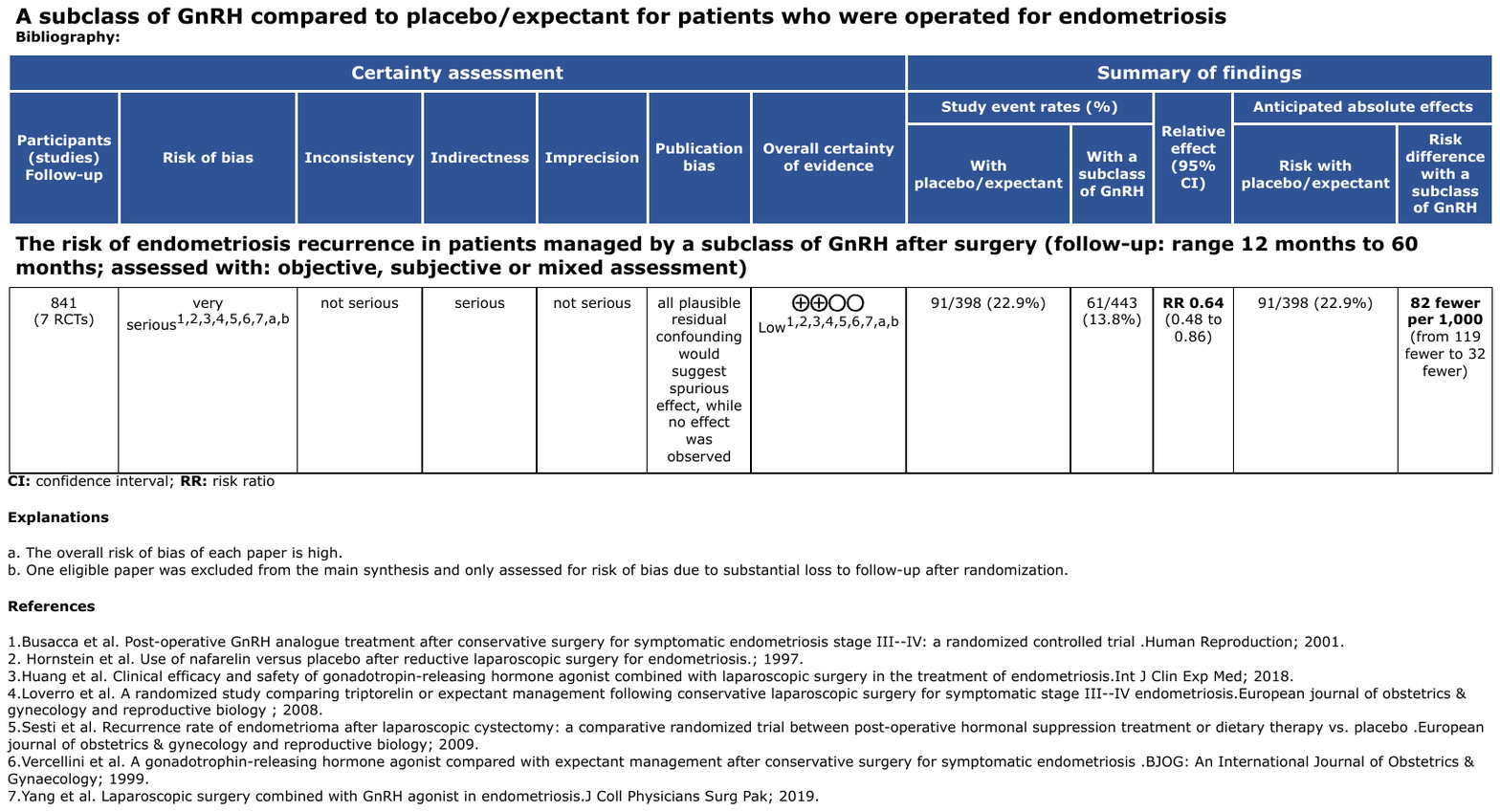}
        \vspace*{-24pt}
        \leftskip -2pt
        \begin{enumerate}[\scriptsize a.]
            \setlength{\itemsep}{-2pt}
            \setlength{\rightskip}{-36pt}
            \item[\scriptsize{$\dagger$}] \scriptsize The table was obtained from GRADEpro;\cite{GRADEpro2025}
            \item[\scriptsize{$\ddagger$}] \scriptsize Endometriosis treatment outcome is affected by surgical skills.\cite{Koninckxetal2021} Since the studies were conducted at different times and in different countries, skills might vary, and spurious confounding may exist;
            \item \scriptsize Risk of bias: Downgraded for the prevalent high/unclear risk of bias in included studies;
            \item \scriptsize Risk of bias: Downgraded because one eligible paper was excluded due to substantial loss to follow-up after randomization, but before the study started;
            \item \scriptsize Risk difference: Estimated from the RR by the application automatically. It was estimated that 82 fewer per $1\,000$ patients who were managed by this subclass of \mbox{GnRH-a} after endometriosis surgery would develop recurrence, compared with those who were not on any \mbox{GnRH-a} therapy. The estimated 95\% CI ranged from 119 fewer to 32 fewer per $1\,000$ patients.
        \end{enumerate}
    \end{minipage}
\end{sideways}
\clearpage

\end{document}